\documentclass[journal]{IEEEtran}

\ifCLASSINFOpdf
\else
   \usepackage[dvips]{graphicx}
\fi
\usepackage{url}

\usepackage{amsmath}
\usepackage{amsfonts}
\usepackage{graphicx}
\usepackage[colorlinks,
linkcolor = blue,
anchorcolor = blue,
citecolor = blue,
]{hyperref}
\usepackage{multirow}
\usepackage{pifont}
\usepackage{multicol}
\usepackage{makecell}
\usepackage{times}
\usepackage{epsfig}
\usepackage{multirow}
\usepackage{overpic}
\usepackage{marvosym}
\usepackage[capitalize]{cleveref}
\usepackage{xcolor}
\usepackage{xspace}
\usepackage{amsmath,amssymb}
\usepackage[accsupp]{axessibility}

\newcommand{\ie}{{\emph{i.e.}}\xspace}

\newcommand{\eg}{{\emph{e.g.}}\xspace}

\crefname{section}{Sec.}{Secs.}
\Crefname{section}{Section}{Sections}
\Crefname{table}{Table}{Tables}
\crefname{table}{Tab.}{Tabs.}

\begin{document}

\title{Super-Resolving Blurry Images with Events}
\author{Chi Zhang, Mingyuan Lin, Xiang Zhang, Chenxu Jiang, Lei Yu
        
\thanks{This work was partially supported by the National Natural Science Foundation of China under Grants 62271354 and 61871297.}
\thanks{C. Zhang, M. Lin, C. Jiang, and L. Yu are with the School of Electronic Information, Wuhan University, Wuhan 430072. X. Zhang is with the Department of Computer Science, ETH Zurich, Switzerland.} 
\thanks{Corresponding author: Lei Yu (ly.wd@whu.edu.cn).}
}

\markboth{IEEE SIGNAL PROCESSING LETTERS}
{Shell \MakeLowercase{\textit{et al.}}: Bare Demo of IEEEtran.cls for IEEE Journals}
\maketitle

\begin{abstract}
Super-resolution from motion-blurred images poses a significant challenge due to the combined effects of motion blur and low spatial resolution. To address this challenge, this paper introduces an {\it E}vent-based {\it B}lurry {\it S}uper {\it R}esolution {\it Net}work (EBSR-Net), which leverages the high temporal resolution of events to mitigate motion blur and improve high-resolution image prediction. Specifically, we propose a multi-scale center-surround event representation to fully capture motion and texture information inherent in events. Additionally, we design a symmetric cross-modal attention module to fully exploit the complementarity between blurry images and events. Furthermore, we introduce an intermodal residual group composed of several residual dense Swin Transformer blocks, each incorporating multiple Swin Transformer layers and a residual connection, to extract global context and facilitate inter-block feature aggregation. Extensive experiments show that our method compares favorably against state-of-the-art approaches and achieves remarkable performance.  
\end{abstract}


\IEEEpeerreviewmaketitle

\section{Introduction}

\IEEEPARstart{M}{otion} blurs often lead to significant performance degradation in Super Resolution (SR) tasks, characterized by motion ambiguities and texture erasure, which poses a substantial challenge for downstream tasks, \eg, autonomous driving~\cite{Wang2022Multitask}, visual detection and tracking~\cite{XIN2024shot,Wu2024Enhanced}, and visual SLAM~\cite{WU0223Improving,Ge2024PIPO_SLAM}. 

While promising progress has been reported in image SR over the past decade~\cite{liang2021swinir,chen2022cross,chen2023dual,park2017joint}, few studies have addressed scenarios involving blurry textures and diverse motion patterns. Consequently, they often lose effectiveness when handling motion-blurred images in real-world dynamic scenarios. Despite decades of separate investigation into the problems of image SR~\cite{liang2021swinir,chen2022cross,chen2023dual} and motion deblurring~\cite{HAN2023MPDNet,Jung2021Multi}, each yielding promising results, simply integrating a motion deblurring module into an image SR architecture may either exacerbate artifacts or compromise detailed information~\cite{singh2014super,zhang2018learning} due to cascading errors. Compared to traditional cascading methods, recent advancements in single-image SR from motion-blurred images have revealed that simultaneous resolution of motion ambiguities can significantly enhance effectiveness~\cite{fang2022high,liang2021flow,niu2021blind}, despite the inherent ill-posed nature of this task~\cite{nah2021ntire}. While kernel-based methods have shown promise in addressing motion-blurred image SR under the assumption of uniform motion~\cite{zhang2018learning,pan2021deep,yun2024kernel}, real-world scenarios often present non-uniform motions, such as those involving non-rigid or moving objects, challenging this assumption. To tackle this issue, various strategies have emerged, including motion flow estimation from video sequences~\cite{park2017joint,bai2024self} and the use of end-to-end deep neural networks~\cite{li2023learning,chen2021scene,li2022face,zhang2020joint,barman2023deep}. However, these approaches are often specialized for specific domains, like faces~\cite{li2023learning,chen2021scene} or text~\cite{li2022face}, or heavily reliant on the performance of the deblurring submodule~\cite{zhang2020joint,barman2023deep}, limiting their applicability to general image SR tasks involving natural scenes with complex motions.  

Recently, several studies have highlighted the advantages of event cameras in Motion-blurred Image Super-Resolution (MSR) in scenes with complex motions~\cite{wang2020event,han2021evintsr,yu2023learning}. These studies report asynchronous event data with extremely low latency (in the order of $\mu$s), which proves effective in recovering accurate sharp details even under non-linear motions and preserving high-resolution information with its high temporal resolution. However, existing methods often experience performance degradation in more complex scenarios due to limitations imposed by sparse coding~\cite{wang2020event,yu2023learning}, as well as accumulated errors in the multi-stage training procedure~\cite{han2021evintsr}.

In this paper, to address the aforementioned issue, we introduce a novel Event-based Blurry Super Resolution Network (EBSR-Net), a one-stage architecture aimed at directly recovering HR sharp image from an motion-blurred LR image across diverse scenarios. We revisit and formulate the MSR task in~\cref{sec:Problem Formulation}, exploring how events can be leveraged to mitigate the ill-posed problem. In~\cref{sec:overall_architecture}, we first introduce a novel Multi-scale Center-surround Event Representation (MCER) module to fully exploit intra-frame motion information for extracting multi-scale textures embedded in events. Then, a Symmetric Cross-Modal Attention (SCMA) module is presented to effectively attend to cross-modal features for subsequent tasks through symmetric querying between frames and events. Furthermore, we design an Intermodal Residual Group (IRG) module consisting of several residual dense Swin Transformer layers and a residual dense connection to facilitate inter-block feature aggregation and extract global context. Overall, the contributions of this paper are three-fold:

\begin{enumerate}
    \item We propose a novel event-based approach, \ie, EBSR-Net, for the single blurry image SR, harnessing cross-modal information between blurry frames and events to reconstruct HR sharp images across diverse scenarios within a one-stage architecture.
    \item We propose an innovative event representation method, \ie, MCER, which comprehensively captures intra-frame motion information through a multi-scale center-surround structure in the temporal domain.
    \item We employ SCMA and IGR modules to achieve effective image restoration by symmetrically querying multimodal feature and facilitating inter-block feature aggregation. 
\end{enumerate}

\section{Methods}

\subsection{Problem Formulation}\label{sec:Problem Formulation}
Due to the imperfection of image sensors, the captured image $B$ may suffer from non-negligible quality degeneration including motion blur and low spatial resolution, which can be related to the high-quality (sharp and high spatial resolution) latent image $L$ as follows:
\begin{equation}\label{eq:blurry_lr}
\begin{split}
    B &= \frac{1}{T} \int_{t\in \mathcal{T}} I(t)dt, \\ I &= D^{\downarrow}(L),
\end{split}
\end{equation}
where $\mathcal{T}\triangleq [0,T]$ denotes the exposure interval of $B$, $T$ is the duration of $\mathcal{T}$, and $D^{\downarrow}(\cdot)$ represents the down-sampled operator to obtain Low Resolution (LR) sharp image $I$ from $L$. Thus the Motion-blurred image Super Resolution (MSR) can be formulated as:
\begin{equation}\label{eq:MSR}
    L = \operatorname{MSR}(B).
\end{equation}
It is obvious that the task of recovering a High-Resolution (HR) sharp image $L$ from a single LR blurry image $B$ poses a severe ill-posed problem. While significant progress has been reported in MSR techniques~\cite{zhang2018learning,pan2021deep,li2022face,li2022face}, these approaches often are specialized for specific domains (\eg, face or text) or rely heavily on the strong assumption of uniform motion, thereby limiting their applicability in natural scenarios with complex motions.

Recently, many methods~\cite{wang2020event,yu2023learning,han2021evintsr} utilize events to tackle MSR task due to their low latency. These events are triggered whenever the log-scale brightness change exceeds the event threshold $c>0$, \ie,
\begin{equation}\label{latent_expEvent}
    \operatorname{log}(I(t,\mathbf{x})) - \operatorname{log}(I(\tau,\mathbf{x})) = p\cdot c,
\end{equation}
where $I(t,\mathbf{x})$ and $I(\tau,\mathbf{x})$ denote the instantaneous intensity at time $t$ and $\tau$ at the pixel position $\mathbf{x}$, and polarity $p\in\{+1,-1\}$ indicates the direction of brightness changes. Hence, the Event-based MSR (E-MSR) task can be represented as:
\begin{equation}\label{eq:EMSR}
        L = \operatorname{E-MSR}(B,\mathcal{E}_{\mathcal{T}}),
\end{equation}
where $\mathcal{E}_{\mathcal{T}}\triangleq {(\mathbf{x}_i,p_i,t_i)}_{t_i\in \mathcal{T}}$ denotes the emitted event stream during the exposure interval of $B$. However, existing approaches~\cite{yu2023learning,han2021evintsr} often suffer performance degradation on more complex scenarios owing to the limitations imposed by sparse coding~\cite{yu2023learning}, and accumulated errors in the multi-stage training procedure~\cite{han2021evintsr}. To address these problems, we introduce a novel Event-based Blurry Super Resolution Network (EBSR-Net), a one-stage architecture aimed at directly recovering $L$ from $B$ and concurrent event stream $\mathcal{E}_{\mathcal{T}}$ across various challenging scenarios.   

\begin{figure}[t]
    \centering
    \includegraphics[width=0.99\linewidth]{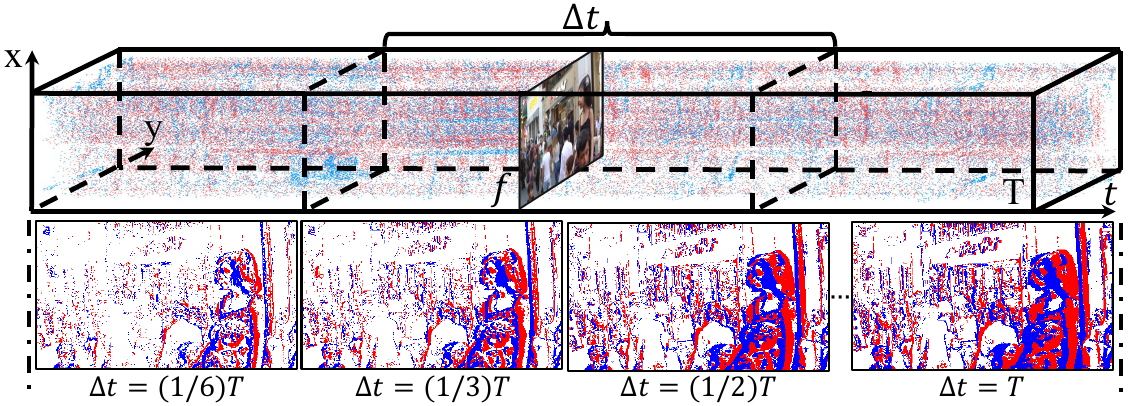}
    \vspace{-5mm}
    \caption{(a) illustrates the details of the proposed Multi-scale Center-surround Event Representation (MCER). Here, $\Delta t$ controls the exposure interval, determining the period utilized for quantizing the event representation.}
    \label{fig:MCER}
    \vspace{-5mm}
\end{figure}

\subsection{Overall Architecture}\label{sec:overall_architecture}

\begin{figure*}[t]
    \centering
    \includegraphics[width=0.92\linewidth]{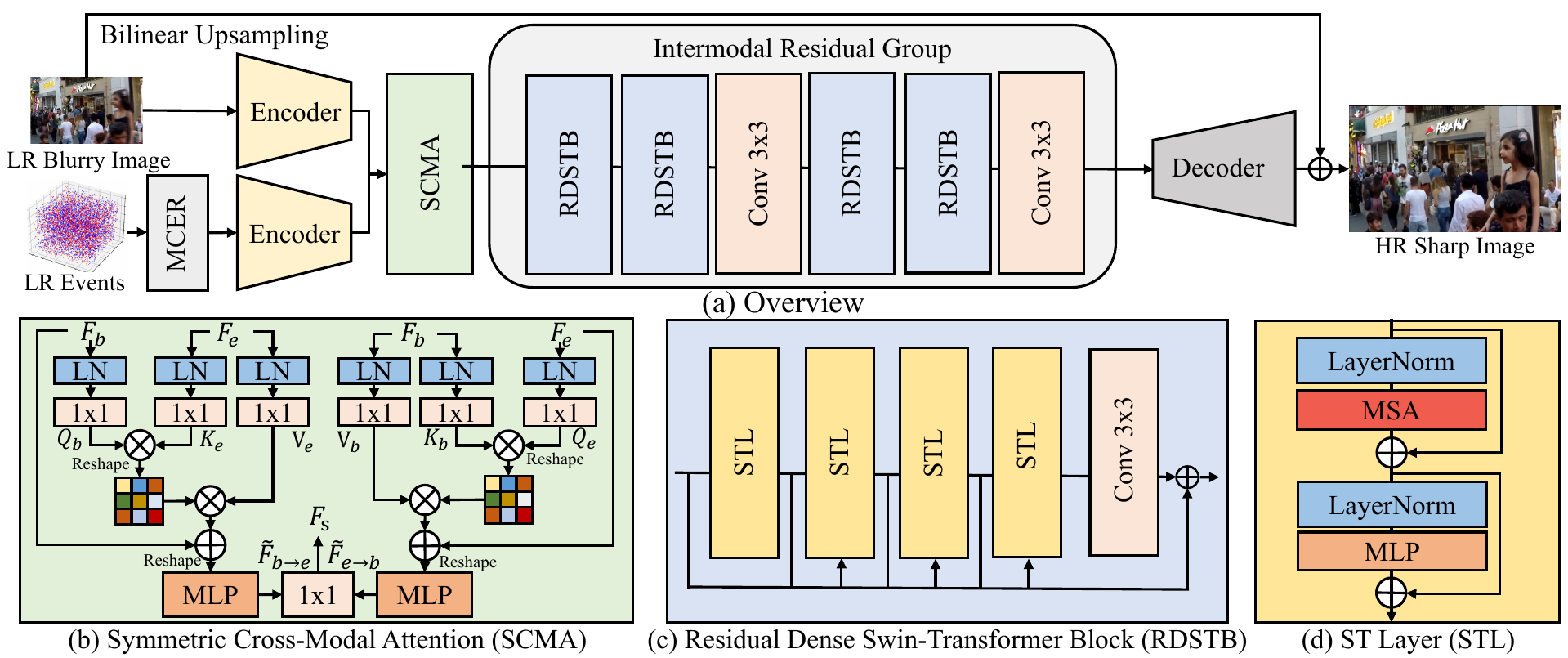}
    \vspace{-1em}
    \caption{(a) illustrates the overview of our proposed EBSR-Net. (b) and (c) provide details of the Symmetric Cross-Modal Attention (SCMA) module and the Residual Dense Swin Transformer Block (RDSTB), respectively. (d) presents the details of the Swin Transformer Layer (STL).}
    \label{fig:overall}
    \vspace{-5mm}
\end{figure*}

The overall architecture of our proposed Event-based Blurry Super Resolution Network (EBSR-Net) is illustrated in~\cref{fig:overall} (a), which is a deep network mainly consisting of a Multi-scale Center-surround Event Representation (MCER) module, a Symmetric Cross-Modal Attention (SCMA) module, and an Intermodal Residual Group (IRG) module. 

\noindent\textbf{Multi-scale Center-surround Event Representation.} The blur degree of a motion-blurred image often varies significantly due to the diverse speeds of the camera or object motion during exposure. To ensure robustness across various scenes with complex motion, we introduce the novel Multi-scale Center-surround Event Representation (MCER) module. This module comprehensively captures intra-frame motion at multiple temporal scales, thereby strengthening the resilience of the event representation. Specifically, we encode event streams $\mathcal{E}_{\mathcal{T}}$ into window-dependent representation frames $E_{\Delta t}(f)$, denoted as
\begin{align}
     E_{\Delta t}(f) &= \operatorname{MCER}(\mathcal{E}_{\mathcal{N}_{\Delta{t}}(f)}), \quad \text{with} \label{blurry_latent_event} 
    \\
    \mathcal{N}_{\Delta{t}}(f) &\triangleq \{t^\prime\mid\left|f-t^\prime\right|\le{\frac{\Delta{t}}{2}},\forall t^\prime\in{\mathcal{T}}\}, \label{multiple_interval} 
\end{align}
where $f$ represents the middle point of the exposure time $T$, $\Delta t$ determines the length of the interval. According to~\cref{blurry_latent_event,multiple_interval}, the event representation results with different $\Delta t$ encode motion information across multiple temporal scales. Additionally, we employ Event Count Map and Timesurface approaches~\cite{xu2021motion} to quantize intra-frame motion information.

\noindent\textbf{Symmetric Cross-Modal Attention.} Compared to existing simple fusion methods between frames and events~\cite{han2021evintsr,xu2021motion}, our proposed Symmetric Cross-Modal Attention (SCMA) module fully exploit the complementary characteristics of multimodal information. Through symmetric querying of multimodal data, it extract adaptively enhanced features for subsequent tasks. As illustrated in~\cref{fig:overall} (b), SCMA module takes the feature of blurry image $F_b$ and events $F_e$ as inputs, yielding symmetric fusion features $F_s$. These features are obtained by encoders consisting of conventional convolutional layers and receiving blurry image $B$ and event map $E_{\Delta t}$.

Unlike conventional self-attention blocks that typically compute queries, keys, and values exclusively from either the frame or event branch of the network, our SCMA leverages multimodal information between frames and events. Queries are calculated from both images and events, with keys and values obtained from the opposite modality. SCMA consists of two parallel self-attention structures and a combination operation, formulated as: 
\begin{equation}\label{eq:sym_attention}
\begin{split}
\operatorname{Att(Q_b,K_e,V_e)} &= V_{e}\operatorname{Softmax}(\frac{Q_{b}^{'}K_e}{\sqrt{d_k}}), 
    \\ \operatorname{Att(Q_e,K_b,V_b)} &= V_{b}\operatorname{Softmax}(\frac{Q_{e}^{'}K_b}{\sqrt{d_k}}),
\end{split}
\end{equation}
where $(\cdot)^{'}$ represents the transpose operator. Note that $Q$, $K$, and $V$, are produced through operations involving normalization and 1$\times$1 convolutional layers. Specifically, $Q_b$ and $Q_e$ are derived from $F_b$ and $F_e$ respectively, similar to $K$ and $V$. Additionally, the adaptively symmetric fusion-based output $F_s$ can be calculated by
\begin{equation}\label{eq:conbine}
        F_s = \operatorname{Conv_{1\times1}}([\tilde{F}_{b\rightarrow e},\tilde{F}_{e\rightarrow b}]),
\end{equation}
where $\tilde{F}_{b\rightarrow e}$ and $\tilde{F}_{e\rightarrow b}$ represent intermediate features from the $Q_b$ and $Q_e$ branches, respectively. These features are generated by reshaping the results that combine attention and original features followed by a Multi-Layer Perceptron (MLP) layer.

\noindent\textbf{Intermodal Residual Group.} The SCMA module aims to extract multimodal information but lacks effective deep features for subsequent tasks. To address this, we introduce the Intermodal Residual Group (IRG) module, which fully exploits deep intermodal information through a meticulously designed group of Residual Dense Swin-Transformer Blocks (RDSTB)~\cite{liang2021swinir,liu2021swin}. The IRG module, depicted in~\cref{fig:overall} (a) and (c), comprises four RDSRBs and two 3$\times$3 convolutional layers. Each RDSRB module contains four residual dense blocks of Swin Transformer Layer (STL). Explicitly, the first RDSTB structure in the IRG module is formulated as:
\begin{equation}\label{eq:STL}
        F_n(d) = \operatorname{STL}([F_n(d-1),F_n(d-2),...,F_s)]),
\end{equation}
where $F_i$ represents the intermediate feature maps of RDSTBs, $d$ denotes the number of STL layers, and $\operatorname{STL}$~\cite{liu2021swin} is based on the standard multi-head self-attention and the original transformer layer~\cite{vaswani2017attention}. Additionally, the output $F_r$ of the first RDSRB module can be obtained by
\begin{equation}\label{eq:Conv_plus}
        F_r = \operatorname{Conv_{3\times3}}(F_n(4))+F_s.
\end{equation}

The final result $F_i$ of the IRG module is computed using three additional RDSTBs and two 3$\times$3 convolutional layers. Consequently, the recovery HR sharp image $\bar{L}$ can be estimated by
\begin{equation}\label{eq:decoder_plus}
        \bar{L} = \operatorname{Dec}(F_r)+\operatorname{BL}(B),
\end{equation}
where $\operatorname{Dec}$ denotes a decoder operator, and $\operatorname{BL}$ represents the bilinear upsampling operation.

\subsection{Loss Function}\label{sec:loss_function}
We supervise the overall architecture by using $L_1$ loss and perceptual similarity loss $\mathcal{L}_{per}$~\cite{zhang2018unreasonable} for better visual quality, which can be formulated as
\begin{equation}\label{eq:totalloss}
    \mathcal{L}_{\text{total}} = \alpha{\| \bar{L} - L\|}_{1} + \beta\mathcal{L}_{\text{per}}(\bar{L}, L),
\end{equation}
where $\alpha$ and $\beta$ are the balancing parameters.

\section{Experiment}\label{Sec:Experiment}

\begin{figure*}[t]
    \centering
    \includegraphics[width=0.99\linewidth]{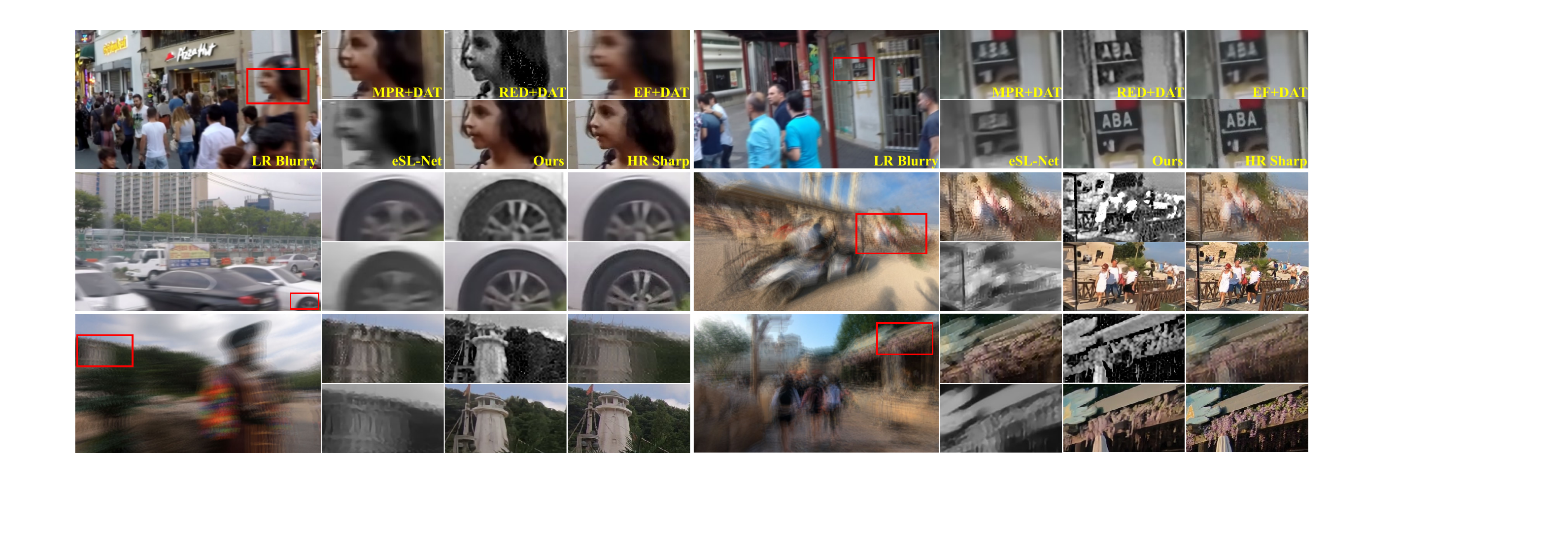}
    \vspace{-1em}
    \caption{Qualitative comparisons of our EBSR-Net with the state-of-the-art methods on the GoPro and the REDS datasets. Six samples are arranged from left to right and top to bottom, with the first three samples from the GoPro dataset and the last three from the REDS dataset.}
    \label{fig:qualitative_comparisons} 
\end{figure*}

\subsection{Datasets and Implementation}\label{sec:datasets and implementation}
The proposed EBSR-Net model, implemented using PyTorch on an NVIDIA GeForce RTX 3090, is trained separately on the training sets of both the GoPro~\cite{nah2017deep} and REDS~\cite{gopro_nah2019ntire} datasets, and we use simulated events and blurry image~\cite{xu2021motion}. Subsequently, evaluations are conducted separately on the testing sets of these two datasets. Furthermore, we use the ADAM optimizer~\cite{kingma2014adam} with an initial learning rate of $10^{-4}$, and the exponential term decays by $0.98$ for every $5$ epoch. The weighting factors $\alpha$ and $\beta$ in \cref{eq:totalloss} are set to 1 and 0.1, respectively. We use Structural SIMilarity (SSIM)~\cite{wang2004image} and Peak Signal to Noise Ratio (PSNR) as performance metrics.

\subsection{Quantitative and Qualitative Evaluation}\label{Quantitative Evaluation} We compare our EBSR-Net to the state-of-the-art Motion Deblurring (MD) methods including MPR~\cite{Zamir2021MPRNet}, RED~\cite{xu2021motion}, and EF~\cite{sun2022event}, and Image Super-Resolution (ISR) approaches, including SwIR~\cite{liang2021swinir}, CAT~\cite{chen2022cross}, and DAT~\cite{chen2023dual}, on both GoPro~\cite{nah2017deep} and REDS~\cite{gopro_nah2019ntire} datasets. These comparison methods are categorized into two stages: MD methods are evaluated first, followed by the ISR methods, denoted as MPR+DAT, and so on. Additionally, a one-stage architecture, \eg, eSL-Net~\cite{wang2020event}, is directly utilized for comparison by utilizing its official code with default parameters. According to the quantitative results shown in~\cref{tab:cznet_basedline_comparison}, our EFSR-Net outperforms state-of-the-art methods by a large margin, achieving an average improvement of 4.12/5.70 dB and 0.0788/0.1607 in PSNR and SSIM on the GoPro and REDS datasets respectively. Meanwhile, our one-stage model only contains 7.3M network parameters, which is much smaller than other methods except eSL-Net. Note that eSL-Net requires 122.8G FLOPs to infer a 160$\times$320 image due to its recursive structure, while our model only needs 41.2G FLOPs, maintaining the overall efficiency. 

\begin{table}[!t]
\centering
\scriptsize
\vspace{-6mm}
\renewcommand{\arraystretch}{1.1}
\caption{Quantitative comparisons on the GoPro~\cite{nah2017deep} and REDS~\cite{gopro_nah2019ntire} datasets. \textbf{Bold} and \underline{Underlined} numbers represent the best and second performance respectively, and OS denotes the One Stage (OS) methods.}
\vspace{-1mm}
\begin{tabular}{lccccccccccc}
\hline
\multirow{2}{*}{Methods} & GoPro~\cite{nah2017deep} & REDS~\cite{gopro_nah2019ntire} & \multirow{2}{*}{\#Param.} & \multirow{2}{*}{OS} & \multirow{2}{*}{Events} \\
& PSNR$\uparrow$/SSIM$\uparrow$ & PSNR$\uparrow$/SSIM$\uparrow$ &
\\ 
\hline
MPR+DAT & 26.86/0.8237  &  19.48/0.5336 & 34.9M  & \ding{55} & \ding{55} 
\\
MPR+CAT & 27.03/0.8266  &  19.62/0.5381 & 36.7M  & \ding{55} & \ding{55} 
\\
MPR+SwIR & 26.87/0.8229  &  19.57/0.5354 & 32.0M  & \ding{55} & \ding{55} 
\\
RED+DAT & 26.91/0.8164 & 21.67/0.5998  &  24.5M  & \ding{55} & \checkmark \\
RED+CAT & 26.74/0.8166 & 21.62/0.5989  &  26.3M  & \ding{55} & \checkmark \\
RED+SwIR & 26.69/0.8134 & 21.58/0.5992  &  21.6M  & \ding{55} & \checkmark \\
EF+DAT & 26.88/0.8262 & \underline{22.75}/\underline{0.7043}  &  23.3M  & \ding{55} & \checkmark \\
EF+CAT & \underline{27.02}/\underline{0.8285} & 21.50/0.6931  &  25.1M  & \ding{55} & \checkmark \\
EF+SwIR & 26.81/0.8253 & 21.49/0.6806  &  20.4M  & \ding{55} & \checkmark \\
eSL-Net & 26.01/0.7818 & 19.82/0.5386  &  \textbf{0.1M}  & \checkmark & \checkmark
\\ \hline
EBSR-Net & \textbf{30.90}/\textbf{0.8969} & \textbf{26.61}/\textbf{0.7629}  &  \underline{7.3M}  & \checkmark & \checkmark
\\ \hline
\end{tabular}
\label{tab:cznet_basedline_comparison}
\vspace{-6mm}
\end{table}

The qualitative comparisons are shown in~\cref{fig:qualitative_comparisons}. The results estimated by the two-stage cascade architecture, \eg, RED~\cite{xu2021motion}+DAT~\cite{chen2023dual}, suffer artifacts and distortions owing to the accumulated errors, leading to significant degradation of the overall quality. Furthermore, one-stage methods like eSL-Net~\cite{wang2020event} exhibit performance degradation in complex scenarios, attributed to limitations imposed by sparse coding. In contrast, our EBSR-Net achieves accurate reconstructions closely resembling the HR ground-truth sharp images.

\begin{figure}[t]
    \vspace{-2mm}
    \centering
    \includegraphics[width=0.99\linewidth]{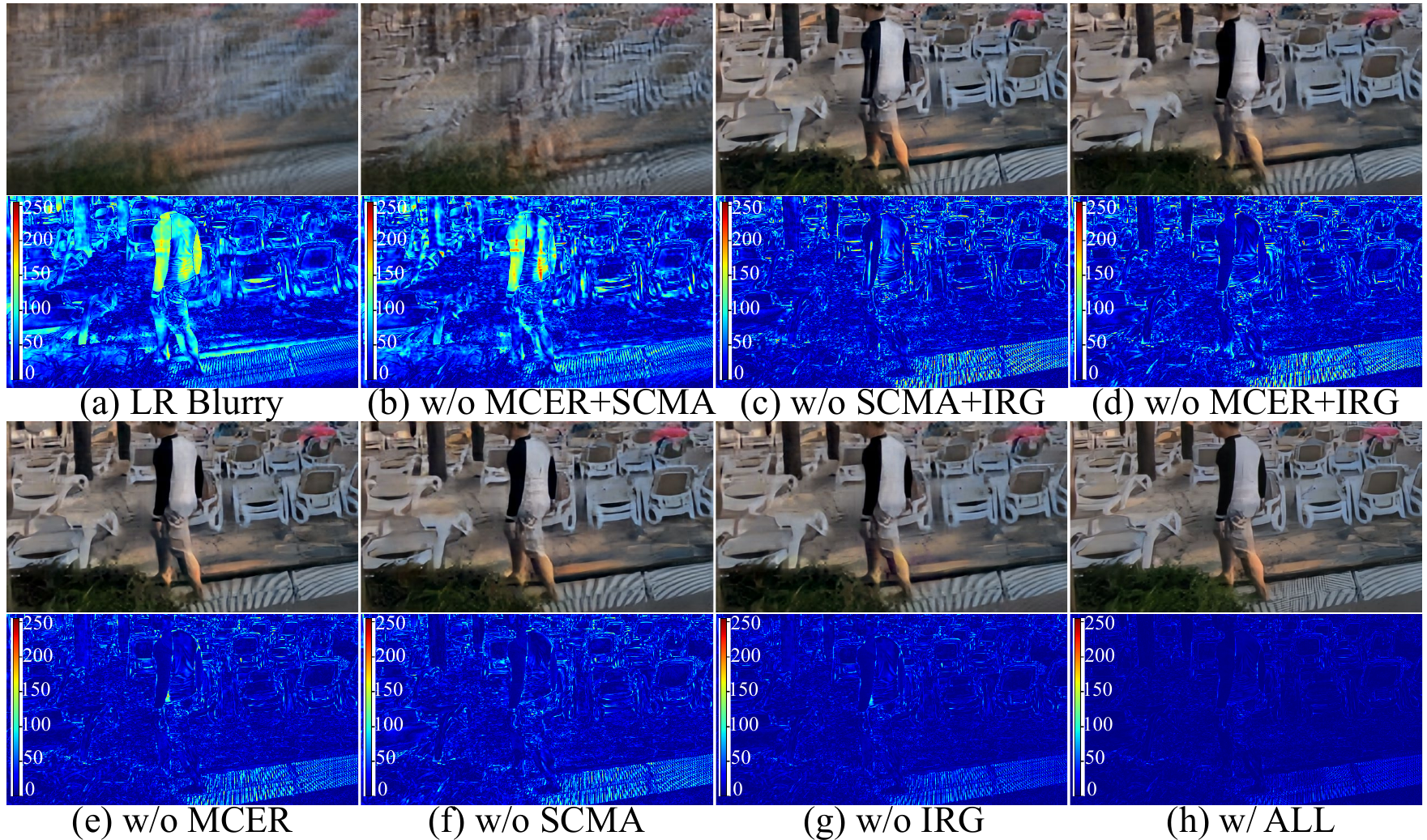}
    \vspace{-3mm}
    \caption{Qualitative ablations of each module of EBSR-Net.}
    \label{fig:ablation}
    \vspace{-5mm}
\end{figure}

\begin{table}[hbpt]
    \centering
    \vspace{-4mm}
    \caption{The Ablation Experimental Results of our EBSR-Net.}
    \vspace{-3mm}
    \resizebox{\linewidth}{!}{
    \begin{tabular}{c|cccccccccc}
\hline
        \multirow{2}{*}{Models} & \multirow{2}{*}{MCER} & \multirow{2}{*}{SCMA} & \multirow{2}{*}{IRG} & \multicolumn{2}{c}{GoPro~\cite{nah2017deep}} &
        \multicolumn{2}{c}{REDS~\cite{gopro_nah2019ntire}}
        \\ \cline{5-6} \cline{7-8}
                &  &  &  & PSNR$\uparrow$ & SSIM$\uparrow$ & PSNR$\uparrow$ & SSIM$\uparrow$ \\ \hline
            \#0    &  \checkmark      &                   &                   &   27.07    &   0.8165    &  24.04 &   0.6756      \\ 
             \#1    &  &    \checkmark        &                   &       27.66 &  0.8336     & 23.87 &   0.6725   \\
             \#2    &  &                   &        \checkmark           &   27.42  &    0.8247   &  22.45 &   0.6232  \\
          \#3 &\checkmark       &      \checkmark             &                   &  \underline{29.67}    &  \underline{0.8810}     & \underline{25.02} &   \underline{0.7181}     \\
           \#4   &\checkmark    &                   &     \checkmark              &  27.79   &  0.8433     &  24.07 &   0.6810 \\
           \#5   &    &          \checkmark         &         \checkmark          &    28.07   &  0.8486     &  24.34 &   0.6992  \\
           \#6 &   \checkmark   &         \checkmark          &       \checkmark            &    \textbf{30.90}    &   \textbf{0.8969}     &   \textbf{26.61}  &   \textbf{0.7629} \\ \hline
        
    \end{tabular}
    }
    \label{tab:ablation_modules}
    \vspace{-6mm}
\end{table}

\subsection{Ablation Study}\label{Ablation Study}
In order to verify the effectiveness of the key components in our EBSR-Net, ablation experiments of the MCER, SCMA, and IRG modules are conducted, as shown in~\cref{tab:ablation_modules}. All models are trained on the same experimental environment and equipment, and we replace the modules with corresponding convolutional layers for a fair comparison. Specifically, removal of the MCER (Case 1, 2, 5), SCMA (Case 0, 2, 4), and IRD (Case 0, 1, 3) modules respectively lead to a degradation of 3.12/3.28/2.91 dB in PSRN and 0.0795/0.0858/0.0722 in SSIM. Furthermore, comparing (b) with (e) and (f) with (h), (c) with (f) and (g) with (h), and (d) with (g) and (e) with (h) demonstrate that EBSR-Net with the SCMA, IRG, and MCER modules respectively give sharper results than network without them. The improvement in both quantitative and qualitative results validates the effectiveness of the proposed three modules.

\section{Conclusion}
In this letter, we present EBSR-Net, an event-based one-stage architecture for recovering HR sharp images from the LR blurry images. We introduce an innovative event representation method, \ie, MCER, which comprehensively captures intra-frame motion information through a multi-scale center-surround structure in the temporal domain. The SCMA and IGR modules are presented to achieve effective image restoration by symmetrically querying multimodal feature and facilitating inter-block feature aggregation. Extensive experimental results show that our method compares favorably against state-of-the-art methods and achieves remarkable performance.  

\bibliographystyle{IEEEtran}
\IEEEtriggeratref{17}
\bibliography{IEEEfull,egbib}

\begin{thebibliography}{10}
\providecommand{\url}[1]{#1}
\csname url@samestyle\endcsname
\providecommand{\newblock}{\relax}
\providecommand{\bibinfo}[2]{#2}
\providecommand{\BIBentrySTDinterwordspacing}{\spaceskip=0pt\relax}
\providecommand{\BIBentryALTinterwordstretchfactor}{4}
\providecommand{\BIBentryALTinterwordspacing}{\spaceskip=\fontdimen2\font plus
\BIBentryALTinterwordstretchfactor\fontdimen3\font minus \fontdimen4\font\relax}
\providecommand{\BIBforeignlanguage}[2]{{%
\expandafter\ifx\csname l@#1\endcsname\relax
\typeout{** WARNING: IEEEtran.bst: No hyphenation pattern has been}%
\typeout{** loaded for the language `#1'. Using the pattern for}%
\typeout{** the default language instead.}%
\else
\language=\csname l@#1\endcsname
\fi
#2}}
\providecommand{\BIBdecl}{\relax}
\BIBdecl

\bibitem{Wang2022Multitask}
Q.~Wang, T.~Han, Z.~Qin, J.~Gao, and X.~Li, ``Multitask attention network for lane detection and fitting,'' \emph{IEEE TNNLS}, vol.~33, no.~3, pp. 1066--1078, 2022.

\bibitem{XIN2024shot}
Z.~Xin, S.~Chen, T.~Wu, Y.~Shao, W.~Ding, and X.~You, ``Few-shot object detection: Research advances and challenges,'' \emph{Information Fusion}, vol. 107, p. 102307, 2024.

\bibitem{Wu2024Enhanced}
Z.~Wu, J.~Wen, Y.~Xu, J.~Yang, X.~Li, and D.~Zhang, ``Enhanced spatial feature learning for weakly supervised object detection,'' \emph{IEEE TNNLS}, vol.~35, no.~1, pp. 961--972, 2024.

\bibitem{WU0223Improving}
Y.~Wu, L.~Wang, L.~Zhang, Y.~Bai, Y.~Cai, S.~Wang, and Y.~Li, ``Improving autonomous detection in dynamic environments with robust monocular thermal slam system,'' \emph{ISPRS Journal of Photogrammetry and Remote Sensing}, vol. 203, pp. 265--284, 2023.

\bibitem{Ge2024PIPO_SLAM}
Y.~Ge, L.~Zhang, Y.~Wu, and D.~Hu, ``Pipo-slam: Lightweight visual-inertial slam with preintegration merging theory and pose-only descriptions of multiple view geometry,'' \emph{IEEE Transactions on Robotics}, vol.~40, pp. 2046--2059, 2024.

\bibitem{liang2021swinir}
J.~Liang, J.~Cao, G.~Sun, K.~Zhang, L.~Van~Gool, and R.~Timofte, ``Swinir: Image restoration using swin transformer,'' in \emph{ICCV}, 2021, pp. 1833--1844.

\bibitem{chen2022cross}
Z.~Chen, Y.~Zhang, J.~Gu, L.~Kong, X.~Yuan \emph{et~al.}, ``Cross aggregation transformer for image restoration,'' \emph{NeurIPS}, vol.~35, pp. 25\,478--25\,490, 2022.

\bibitem{chen2023dual}
Z.~Chen, Y.~Zhang, J.~Gu, L.~Kong, X.~Yang, and F.~Yu, ``Dual aggregation transformer for image super-resolution,'' in \emph{ICCV}, 2023, pp. 12\,312--12\,321.

\bibitem{park2017joint}
H.~Park and K.~Mu~Lee, ``Joint estimation of camera pose, depth, deblurring, and super-resolution from a blurred image sequence,'' in \emph{ICCV}, 2017, pp. 4613--4621.

\bibitem{HAN2023MPDNet}
G.~Han, M.~Wang, H.~Zhu, and C.~Lin, ``Mpdnet: An underwater image deblurring framework with stepwise feature refinement module,'' \emph{Engineering Applications of Artificial Intelligence}, vol. 126, p. 106822, 2023.

\bibitem{Jung2021Multi}
H.~Jung, Y.~Kim, H.~Jang, N.~Ha, and K.~Sohn, ``Multi-task learning framework for motion estimation and dynamic scene deblurring,'' \emph{IEEE TIP}, vol.~30, pp. 8170--8183, 2021.

\bibitem{singh2014super}
A.~Singh, F.~Porikli, and N.~Ahuja, ``Super-resolving noisy images,'' in \emph{CVPR}, 2014, pp. 2846--2853.

\bibitem{zhang2018learning}
K.~Zhang, W.~Zuo, and L.~Zhang, ``Learning a single convolutional super-resolution network for multiple degradations,'' in \emph{CVPR}, 2018, pp. 3262--3271.

\bibitem{fang2022high}
N.~Fang and Z.~Zhan, ``High-resolution optical flow and frame-recurrent network for video super-resolution and deblurring,'' \emph{Neurocomputing}, vol. 489, pp. 128--138, 2022.

\bibitem{liang2021flow}
J.~Liang, K.~Zhang, S.~Gu, L.~Van~Gool, and R.~Timofte, ``Flow-based kernel prior with application to blind super-resolution,'' in \emph{CVPR}, 2021, pp. 10\,601--10\,610.

\bibitem{niu2021blind}
W.~Niu, K.~Zhang, W.~Luo, and Y.~Zhong, ``Blind motion deblurring super-resolution: When dynamic spatio-temporal learning meets static image understanding,'' \emph{IEEE TIP}, vol.~30, pp. 7101--7111, 2021.

\bibitem{nah2021ntire}
S.~Nah, S.~Son, S.~Lee, R.~Timofte, and K.~M. Lee, ``Ntire 2021 challenge on image deblurring,'' in \emph{CVPR}, 2021, pp. 149--165.

\bibitem{pan2021deep}
J.~Pan, H.~Bai, J.~Dong, J.~Zhang, and J.~Tang, ``Deep blind video super-resolution,'' in \emph{ICCV}, 2021, pp. 4811--4820.

\bibitem{yun2024kernel}
J.-S. Yun, M.~H. Kim, H.-I. Kim, and S.~B. Yoo, ``Kernel adaptive memory network for blind video super-resolution,'' \emph{Expert Systems with Applications}, vol. 238, p. 122252, 2024.

\bibitem{bai2024self}
H.~Bai and J.~Pan, ``Self-supervised deep blind video super-resolution,'' \emph{IEEE TPAMI}, 2024.

\bibitem{li2023learning}
X.~Li, W.~Zuo, and C.~C. Loy, ``Learning generative structure prior for blind text image super-resolution,'' in \emph{CVPR}, 2023, pp. 10\,103--10\,113.

\bibitem{chen2021scene}
J.~Chen, B.~Li, and X.~Xue, ``Scene text telescope: Text-focused scene image super-resolution,'' in \emph{CVPR}, 2021, pp. 12\,026--12\,035.

\bibitem{li2022face}
X.~Li, C.~Chen, X.~Lin, W.~Zuo, and L.~Zhang, ``From face to natural image: Learning real degradation for blind image super-resolution,'' in \emph{ECCV}.\hskip 1em plus 0.5em minus 0.4em\relax Springer, 2022, pp. 376--392.

\bibitem{zhang2020joint}
D.~Zhang, Z.~Liang, and J.~Shao, ``Joint image deblurring and super-resolution with attention dual supervised network,'' \emph{Neurocomputing}, vol. 412, pp. 187--196, 2020.

\bibitem{barman2023deep}
T.~Barman and B.~Deka, ``A deep learning-based joint image super-resolution and deblurring framework,'' \emph{IEEE Transactions on Artificial Intelligence}, 2023.

\bibitem{wang2020event}
B.~Wang, J.~He, L.~Yu, G.-S. Xia, and W.~Yang, ``Event enhanced high-quality image recovery,'' in \emph{ECCV}.\hskip 1em plus 0.5em minus 0.4em\relax Springer, 2020, pp. 155--171.

\bibitem{han2021evintsr}
J.~Han, Y.~Yang, C.~Zhou, C.~Xu, and B.~Shi, ``Evintsr-net: Event guided multiple latent frames reconstruction and super-resolution,'' in \emph{ICCV}, 2021, pp. 4882--4891.

\bibitem{yu2023learning}
L.~Yu, B.~Wang, X.~Zhang, H.~Zhang, W.~Yang, J.~Liu, and G.-S. Xia, ``Learning to super-resolve blurry images with events,'' \emph{IEEE TPAMI}, 2023.

\bibitem{xu2021motion}
F.~Xu, L.~Yu, B.~Wang, W.~Yang, G.-S. Xia, X.~Jia, Z.~Qiao, and J.~Liu, ``Motion deblurring with real events,'' in \emph{ICCV}, 2021, pp. 2583--2592.

\bibitem{liu2021swin}
Z.~Liu, Y.~Lin, Y.~Cao, H.~Hu, Y.~Wei, Z.~Zhang, S.~Lin, and B.~Guo, ``Swin transformer: Hierarchical vision transformer using shifted windows,'' in \emph{ICCV}, 2021, pp. 10\,012--10\,022.

\bibitem{vaswani2017attention}
A.~Vaswani, N.~Shazeer, N.~Parmar, J.~Uszkoreit, L.~Jones, A.~N. Gomez, {\L}.~Kaiser, and I.~Polosukhin, ``Attention is all you need,'' \emph{NeurIPS}, vol.~30, 2017.

\bibitem{zhang2018unreasonable}
R.~Zhang, P.~Isola, A.~A. Efros, E.~Shechtman, and O.~Wang, ``The unreasonable effectiveness of deep features as a perceptual metric,'' in \emph{CVPR}, 2018, pp. 586--595.

\bibitem{nah2017deep}
S.~Nah, T.~Hyun~Kim, and K.~Mu~Lee, ``Deep multi-scale convolutional neural network for dynamic scene deblurring,'' in \emph{CVPR}, 2017, pp. 3883--3891.

\bibitem{gopro_nah2019ntire}
S.~Nah, S.~Baik, S.~Hong, G.~Moon, S.~Son, R.~Timofte, and K.~Mu~Lee, ``Ntire 2019 challenge on video deblurring and super-resolution: Dataset and study,'' in \emph{CVPRW}, 2019, pp. 1974--1984.

\bibitem{kingma2014adam}
D.~P. Kingma and J.~Ba, ``Adam: A method for stochastic optimization,'' in \emph{ICLR}, 2015.

\bibitem{wang2004image}
Z.~Wang, A.~C. Bovik, H.~R. Sheikh, and E.~P. Simoncelli, ``Image quality assessment: from error visibility to structural similarity,'' \emph{IEEE TIP}, vol.~13, no.~4, pp. 600--612, 2004.

\bibitem{Zamir2021MPRNet}
S.~W. Zamir, A.~Arora, S.~Khan, M.~Hayat, F.~S. Khan, M.-H. Yang, and L.~Shao, ``Multi-stage progressive image restoration,'' in \emph{CVPR}, 2021.

\bibitem{sun2022event}
L.~Sun, C.~Sakaridis, J.~Liang, Q.~Jiang, K.~Yang, P.~Sun, Y.~Ye, K.~Wang, and L.~V. Gool, ``Event-based fusion for motion deblurring with cross-modal attention,'' in \emph{ECCV}.\hskip 1em plus 0.5em minus 0.4em\relax Springer, 2022, pp. 412--428.

\end{thebibliography}

\end{document}